# EventFlow: Real-Time Neuromorphic Event-Driven Classification of Two-Phase Boiling Flow Regimes

Sanghyeon Chang[1] | Srikar Arani[2] | Nishant Sai Nuthalapati[2] | Youngjoon Suh[1] | Nicholas Choi[1] | Siavash Khodakarami[3] | Md Rakibul Hasan Roni[3] | Nenad Miljkovic[3,4,5,6,7,8] | Aparna Chandramowlishwaran[2] | Yoonjin Won[1,2,*]

[1]Department of Mechanical and Aerospace Engineering, University of California, Irvine, Irvine, CA, USA

[2]Department of Electrical Engineering and Computer Science, University of California, Irvine, Irvine, CA, USA

[3]Department of Mechanical Science and Engineering, University of Illinois at Urbana-Champaign, Urbana, IL, USA

[4]Department of Electrical and Computer Engineering, University of Illinois at Urbana-Champaign, Urbana, IL, USA

[5]Institute for Sustainability, Energy, and Environment, University of Illinois Urbana-Champaign, Urbana, IL, USA

[6]Materials Research Laboratory, University of Illinois Urbana-Champaign, Urbana, IL, USA

[7]Air Conditioning and Refrigeration Center, University of Illinois Urbana-Champaign, Urbana, IL, USA

[8]International Institute for Carbon Neutral Energy Research (WPI-I2CNER), Kyushu University, 744 Moto-oka, Nishi-ku, Fukuoka 819-0395, Japan

Correspondence

Yoonjin Won, University of California, Irvine 4200 Engineering Gateway, Irvine, California, 92697.

Email: won@uci.edu

## Abstract

Flow boiling is an efficient heat transfer mechanism capable of dissipating high heat loads with minimal temperature variation, making it an ideal thermal management method. However, sudden shifts between flow regimes can disrupt thermal performance and system reliability, highlighting the need for accurate and low-latency real-time monitoring. Conventional optical imaging methods are limited by high computational demands and insufficient temporal resolution, making them inadequate for capturing transient flow behavior. To address this, we propose a real-time framework based on signals from neuromorphic sensors for flow regime classification. Neuromorphic sensors detect changes in brightness at individual pixels, which typically correspond to motion at edges, enabling fast and efficient detection without full-frame reconstruction, providing event-based information. We develop five classification models using both traditional image data and event-based data, demonstrating that models leveraging event data outperform frame-based approaches due to their sensitivity to dynamic flow features. Among these models, the event-based long short-term memory model provides the best balance between accuracy and speed, achieving 97.6% classification accuracy with a processing time of 0.28 ms. Our asynchronous processing pipeline supports continuous, low-latency predictions and delivers stable output through a majority voting mechanisms, enabling reliable real-time feedback for experimental control and intelligent thermal management.

## Keywords

Flow boiling, Computer vision, Neuromorphic event camera, Flow regime classification, Real-time classification.

# Introduction

Two-phase boiling is a compact and highly efficient thermal management approach that utilizes liquid-to-vapor phase change to achieve high heat fluxes.[1,2] This mechanism enables effective heat dissipation in applications ranging from industrial power systems to the cooling of high-performance electronics.[3–6] As thermal demands rise across energy-intensive and climate-sensitive sectors, two-phase boiling has become increasingly important for addressing these challenges.[7,8] Its unique capabilities make it especially relevant for emerging technologies such as artificial intelligence (AI) data centers, renewable energy systems, and next-generation cooling solutions.[9–11]

The increasing demand for high-power-density cooling has intensified interest in flow boiling, a process in which a working fluid circulates through channels, absorbs heat from a heated surface, and undergoes phase change along the flow path.[12,13] This convective motion enhances heat removal from compact regions and allows precise control of temperature gradients, making flow boiling particularly well-suited for compact and high-flux thermal management.[14]

Flow regimes in boiling systems define the specific arrangement and movement of liquid and vapor phases within a channel, directly influencing heat transfer efficiency, system stability, and overall performance.[15] However, the behavior of bubbles in confined, high-velocity environments introduces substantial instability due to their intrinsic complexity. Bubble nucleation, growth, coalescence, and departure are strongly influenced by channel geometry, flow conditions, and surface properties, posing challenges for system design and control.[16–20] Because flow regimes exhibit distinct phase distributions and thermal behaviors, their accurate classification is essential for optimizing thermal management systems and preventing operational inefficiencies.[21,22] This need becomes especially critical in modern applications, where miniaturized channels (< 1 mm) and high heat fluxes (> 100 W/cm$^2$) demand precise control over boiling behavior. As a result, flow regime classification has emerged as a key strategy for capturing the dynamic nature of two-phase flows and enabling effective real-time diagnostics and control.

Several techniques have been explored for flow regime classification, including sensor-based measurements and acoustic analysis.[23–26] While promising, these methods face practical limitations that reduce generalizability and robustness. Pressure, temperature, and mechanical vibration sensors, often used in machine learning models to detect phase transitions, are highly sensitive to specific experimental configurations and boundary conditions, limiting their applicability to broader systems. Impedance probes combined with neural networks have shown promise for gas-liquid flow detection by analyzing space-frequency features from multi-electrode signals.[27] However, while these techniques provide localized measurements, their spatial coverage is limited to the probe volume and may not resolve distributed or rapidly evolving interfacial structures that are critical for distinguishing complex two-phase flow regimes. Acoustic methods can detect boiling instabilities through sound wave patterns but are vulnerable to background noise and signal interference, which can degrade performance in uncontrolled environments.[28]

To address the limitations of conventional sensing approaches, researchers have increasingly adopted optical imaging coupled with deep learning to identify flow regimes from observable bubble shapes and spatial distributions that reflect the underlying boiling physics.[29] Optical imaging directly captures visual features of the flow field, making it less dependent on specific sensor placement or boundary conditions. Unlike point sensors with limited spatial coverage, optical images preserve fine-scale interfacial structures across the entire field of view, which are essential for distinguishing between subtle regime transitions. Moreover, optical methods are inherently resistant to background acoustic noise and mechanical interference, as they rely on visual data rather than physical signal propagation.

Optical imaging, particularly when combined with deep learning-based convolutional neural networks (CNNs), has been widely used for bubble detection and flow regime classification.[30–34] High-speed cameras can capture rapid boiling phenomena with microsecond precision, but their d74ata-intensive outputs significantly increase computational overhead. These methods record every pixel at every timestep, often across multiple color channels, resulting in excessive data redundancy, as many pixels merely capture static backgrounds.[35] Moreover, short recording durations during high-speed imaging and high-power consumption make high-speed cameras impractical for continuous operation in industrial applications. Deep learning methods have attempted to address these challenges using encoder-decoder architectures, achieving competitive classification accuracy.[32] However, such models often suffer from low recall and frequent misclassification, particularly under unstable or transitional flow conditions.

Neuromorphic event cameras present a unique and promising alternative by providing real-time visual sensing solutions that combine low power consumption, minimal data redundancy, and reliable classification.[36,37] They produce sparse data streams with precise timestamps instead of full image frames. This event-based sensing reduces data redundancy, preserves fine temporal features, and eliminates motion blur. Event cameras also operate with high dynamic range and low power consumption, enabling continuous monitoring in challenging environments. This sensing mechanism minimizes data redundancy while maintaining high temporal resolution, making them well-suited for capturing fast, localized variations in dynamic two-phase flows. Their low power consumption and ability to operate continuously make them ideal for real-time operation of thermal management, monitoring applications as well as flow visualization.[38,39] The event-based approach requires optical access and is therefore explicitly limited to environments where optical access is feasible such as transparent sections, inspection windows, or endoscopic views, and it is not intended for sealed or highly miniaturized channels. Still it has the potential to serve as a complementary modality that augments embedded sensors by providing high fidelity spatiotemporal information, improving calibration and drift detection, and strengthening diagnostic robustness and generalizability.

This work suggests event-based sensing for flow boiling classification by integrating neuromorphic vision with deep learning. To overcome the limitations of conventional sensors and frame-based imaging, we develop a classification pipeline using asynchronous event data to capture fast two-phase dynamics. A dataset of synchronized high-speed and event recordings supports multiple deep learning models training and evaluation. Performance is assessed across flow regimes and sensing modalities in terms of accuracy, temporal resolution, and data efficiency. Results show that event-driven approaches enable accurate, real-time classification with reduced computational and power demands, supporting advanced thermal and energy system applications.

Flow regimes in boiling systems describe the spatial and temporal distribution of liquid and vapor phases within a channel. The flow regime is fundamental to understanding two-phase heat transfer behavior.[6] In this study, we consider seven representative flow regimes observed during subcooled-to-saturated flow boiling in horizontal channels: (B) bubbly, (EB) elongated bubbly, (S) slug, (SS) stratified smooth, (SW) stratified wavy, (A) annular, and (U) unstable. These regimes reflect progressive changes in phase distribution and interface dynamics as heat flux increases, ranging from discrete bubble nucleation to vapor-dominated annular flow and transitional instabilities near critical heat flux. Accurate identification of these regimes is essential for understanding flow development and predicting thermal performance under varying operating conditions.

Among multiple flow boiling regimes, (B) bubbly flow features discrete vapor bubbles dispersed in a continuous liquid phase. As the flow develops from bubbly to elongated bubbly regimes (EB), bubbles grow and merge into larger vapor structures that span most of the channel width. The resulting elongated bubbly regime consists of multiple small bubbles that cluster together, creating vapor structures that appear narrow but stretched along the flow direction. Elongated bubbles are still discrete, which makes differs other continuous vapor blanket in such stratified smooth flow. (S) In slug, further vapor accumulation leads to slug flow, which involves alternating vapor plugs and liquid slugs with strong interface oscillations. Unlike elongated bubbly, slug flow shows periodic sequence of vapor–liquid segments along the channel. At higher vapor fractions, the flow becomes stratified, where gravity separates the phases into horizontal layers. (SS) In stratified smooth flow, the liquid forms a stable layer along the lower wall. The interface is flat and continuous without visible disturbances. (SW) In stratified wavy flow, the interface develops surface undulations driven by shear or surface tension effects.[40] These interfacial waves occur while the lower liquid layer remains continuous, distinguishing this regime from annular flow. (A) Annular flow emerges with continued heating, where a high-speed vapor core is surrounded by a thin liquid film coating the channel walls.[41] Unlike stratified flows, the annular film extends circumferentially and is maintained continuously around the channel perimeter, often with interfacial ripples or irregularities. (U) Near critical heat flux, unstable regimes appear with irregular and mixed flow patterns that are difficult to classify and often signal boiling-induced instabilities or dry-out.

A major challenge in flow regime classification is the similarity between certain patterns, particularly stratified smooth, stratified wavy, and annular flows.[42] These regimes share a structure where liquid forms a continuous film along the walls and vapor occupies the center. Subtle differences in waviness and film thickness complicate classification, particularly for optical models that rely on static features. Event-based sensing offers higher temporal precision, enhancing sensitivity to transient flow dynamics and improving distinction between visually similar regimes.

Figure 1a shows a sequence of subcooled inlet flow boiling patterns, progressing from bubbly to dry-out. This captures regime evolution from nucleate boiling to near-critical heat flux, with increasing vapor dominance and phase asymmetry. Due to rapid, unpredictable transitions in flow boiling, there is a pressing need for real-time diagnostic and classification tools to track these changes and enable precise control in high-power-density cooling systems.[43–45]

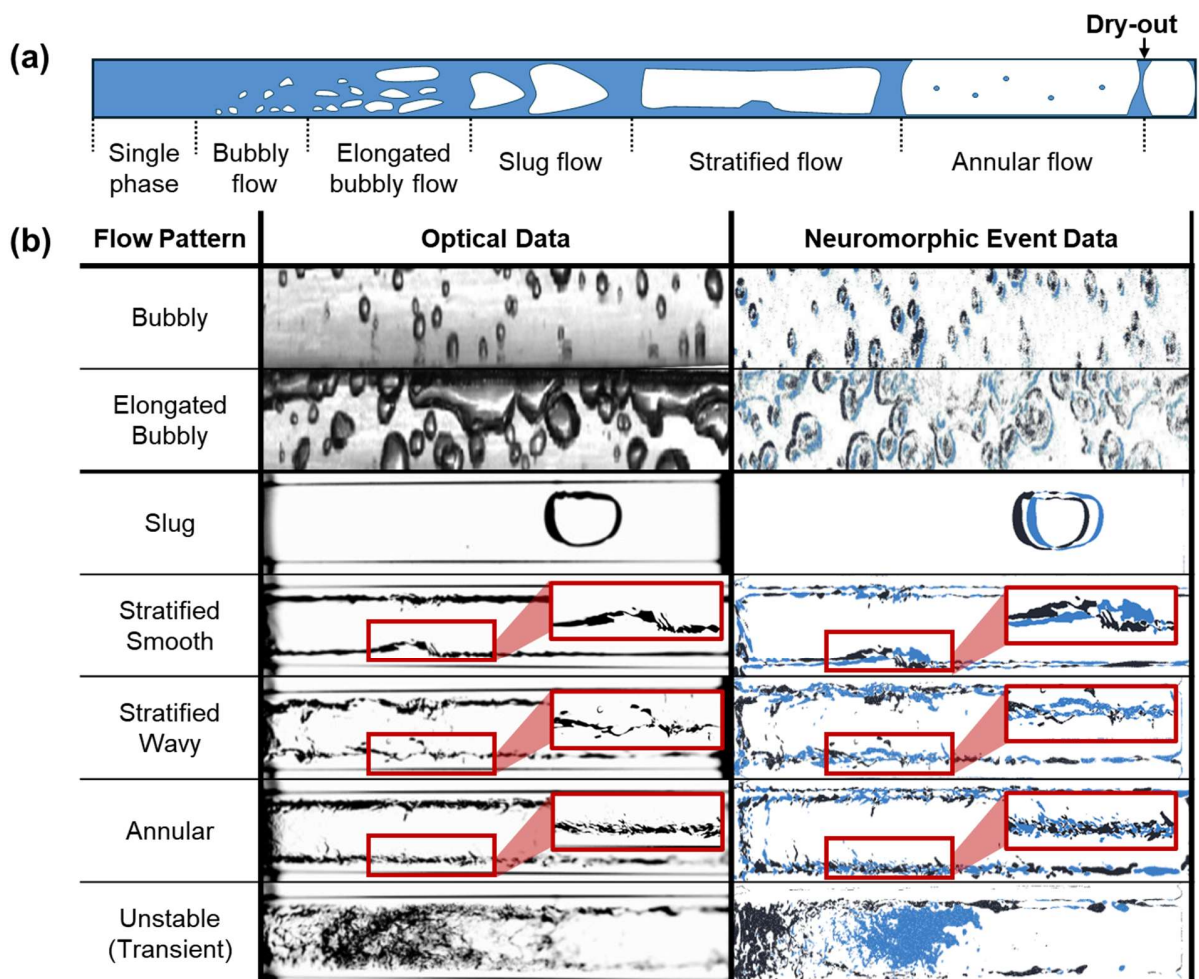

**Figure 1. Visualization of optical and event representation of seven different flow patterns.** (a) Illustration

showing various flow boiling patterns like bubbly, elongated bubbly, slug, stratified and annular flow followed by dry-out. These regimes exhibit increasing vapor dominance and spatial asymmetry with rising thermal load. Schematic not to scale. (b) Side-by-side comparison of optical and neuromorphic event-based images for the seven flow boiling regimes including (B) Bubbly, (EB) Elongated Bubbly, (S) Slug, (SS) Stratified Smooth, (SW) Stratified Wavy, (A) Annular, and (U) Unstable regimes. The neuromorphic mechanism captures only dynamic brightness changes, removing static background and enhancing the visibility of interfacial features. The events data are collected via the use of event camera (B or EB) or synthetic algorithm (S, SS, SW, A, U). The subtle differences in flows like in (SS), (SW), and (A) regimes are more clearly distinguished in the event-based images, as indicated by red rectangles. This highlights the potential for neuromorphic sensing in analyzing fast and visually ambiguous two-phase flow structures.

## Results and Discussion

**Neuromorphic Data**

A neuromorphic event camera (Prophesee EVK3) captures asynchronous, per-pixel brightness changes, generating sparse and high temporal resolution event streams. Unlike conventional cameras that operate on fixed frame intervals, event cameras detect variations in logarithmic intensity and emit events only when a predefined contrast threshold is exceeded. Each event is a four-dimensional tuple ($x$, $y$, $t$, $p$), where $x$ and $y$ are the pixel location, $t$ is the timestamp, and $p$ is the polarity indicating brightness change. Figure 2a illustrates this process by comparing conventional frame-based images with event-based outputs. While conventional images capture full scenes at fixed intervals, event data highlights only the pixels where brightness changes occur. In the event representation, an "on" or "off" event is generated depending on the polarity of the brightness change. This selective triggering allows the camera to ignore static background information and focus entirely on dynamic features, making it highly effective for capturing fast and transient phenomena such as flow boiling.

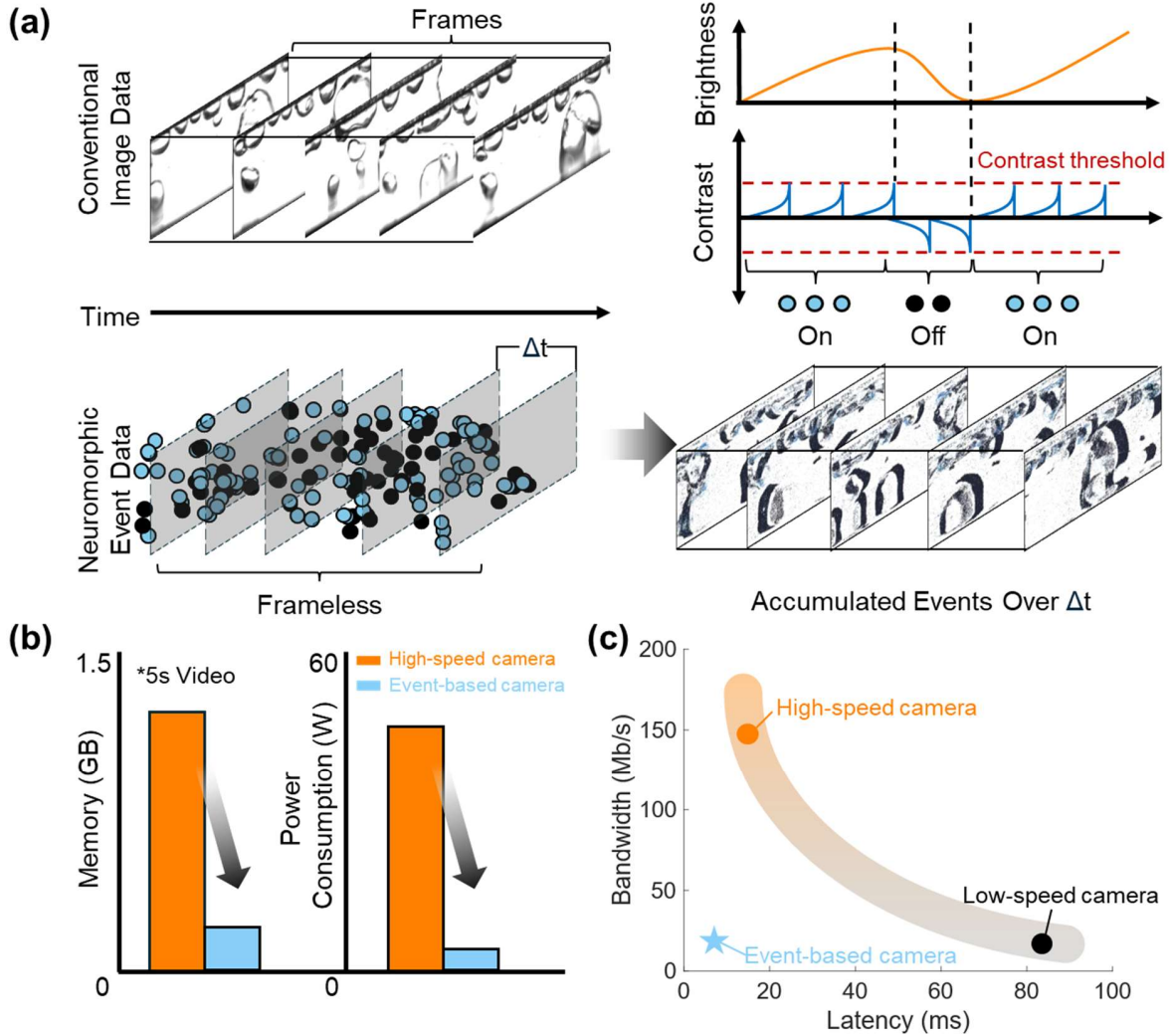

**Figure 2. Comparison of High-Speed and Event Cameras for Flow Boiling Analysis.** (a) Event cameras emit

asynchronous brightness change events at each pixel, capturing motion only and enabling sparse frame reconstruction over a chosen time window (Δt). (b) This sparsity greatly reduces resource use. A 5-second recording at 2,000 frames per second (fps) decreases from roughly 1.2 GB and 55 W (FASTCAM Mini AX100) to about 150 MB and 4.5 W (Prophesee EVK3). (c) Sub-millisecond latency and a 45 fps-equivalent data rates give event cameras responsiveness comparable to a 5,000 fps system while keeping the computational load low.

Figure 1b presents a comparative visualization of optical and event data across seven distinct flow regimes. Optical imaging captures fine spatial details such as textures and phase boundaries, while event-based sensing provides high temporal resolution, revealing rapid transitions and unstable patterns that may be missed by conventional frame-based methods.

By capturing pixel-level brightness changes with microsecond precision, event cameras offer exceptional temporal resolution and eliminate motion blur, enabling accurate tracking of fast, localized dynamics. As shown in Figure 2b, their sparse data stream significantly reduces memory and computation load compared to traditional high-speed imaging. Power consumption is also significantly lower, with high-speed cameras using approximately 55 W and event cameras consuming only 4.5 W, making them ideal for real-time flow instability detection.

Figure 2c shows that event cameras maintain sub-millisecond latency and ultralow bandwidth, matching the responsiveness of a 5,000 fps system while using the bandwidth of only 45 fps.[46] By capturing only meaningful visual changes, they reduce the need for computationally intensive pre- or post-processing, enabling efficient monitoring in flow boiling research.

**Machine Learning Models for Classification**

The proposed framework explores multiple ML architectures, including *Optical CNN*[47], *Event CNN*[47], *Fourier k-NN*[48], *Event SNN*[49], and *Event LSTM*[50]. Each model is tailored to exploit the unique characteristics of optical and neuromorphic event data for classifying flow boiling regimes, as illustrated in Figure 3. This multi-model approach enables a comparative evaluation of frame-based and event-based learning strategies under the same experimental conditions. The classification task focuses on identifying seven distinct flow regimes, each exhibiting unique interfacial structures, spatial textures, and temporal behaviors. By leveraging complementary representations of the boiling process, including dense frame sequences from high-speed optical imaging and sparse high-frequency event streams from neuromorphic sensing, the framework aims to determine which data modality and model architecture are most effective for real-time regime recognition.

*Optical CNN* refers to a convolutional neural network architecture applied to high-speed optical image sequences.[47] Optical frames are manually annotated based on observed flow characteristics and categorized into distinct flow regimes. This labeled dataset is used to train the model to recognize interfacial structures and dynamic patterns associated with each regime. The model consists of stacked convolutional layers followed by pooling, activation, and fully connected layers, which extract spatial features from dense visual inputs. It processes individual frames or short sequences to identify textures and structural features such as vapor–liquid interfaces, bubble contours, and flow boundaries. Feature maps are progressively downsampled through convolution and pooling layers, and the resulting high-level spatial features are passed to fully connected layers that produce a flow regime classification.

*Event CNN* refers to a convolutional neural network architecture[47] applied to reconstructed images generated from accumulated event data. While the original event stream is asynchronous and sparse, this transformation produces image representations that preserve temporal information and enable spatial feature extraction using standard frame-based models. As illustrated in Figure 3c, the *Event CNN* architecture follows a standard CNN pipeline, processing spatially and temporally sparse inputs to extract meaningful spatiotemporal features that characterize distinct flow patterns.

The Fourier k-nearest neighbors (*k-NN*) method refers to a classification pipeline that operates on event frames generated from accumulated neuromorphic events. Each accumulated frame is transformed into the frequency domain using the Fast Fourier Transform (FFT) to extract dominant spatial frequency components. These spectral features are then classified using the k-NN algorithm based on similarity in the frequency space.[48] This approach focuses on capturing periodic flow characteristics rather than localized spatial features, offering an alternative representation for distinguishing flow regimes.

Event spiking neural network (*Event SNN*) refers to a spiking neural network architecture that directly consumes raw neuromorphic event streams without frame conversion.[49] The model encodes asynchronous events as spike trains and processes them using spiking neurons and synaptic connections that mimic biological neural systems.[51] The network consists of multiple layers of spiking units, each updating its membrane potential and firing only in response to significant input activity. The spatiotemporal features of the event stream are captured through temporal coding mechanisms, and the spike-based output is processed through a readout layer to assign flow regime classes. The architecture operates in an event-driven manner, maintaining high temporal precision and computational sparsity.

Event long short-term memory (*Event LSTM*) refers to a recurrent neural network architecture designed to model the temporal evolution of boiling behavior using raw neuromorphic event streams.[50] Instead of using accumulated frames or reconstructed images, the model processes raw asynchronous event data. Events are filtered by region of interest and time, then organized into temporally ordered sequences labeled using synchronized optical annotations. LSTM layers capture short- and long-term dependencies by updating internal states over time, enabling detection of dynamic features such as bubble nucleation and flow instabilities. The final hidden state is passed to a fully connected layer for flow regime classification, leveraging the temporal precision and sparsity of event-based sensing to model complex boiling dynamics directly from the event stream.

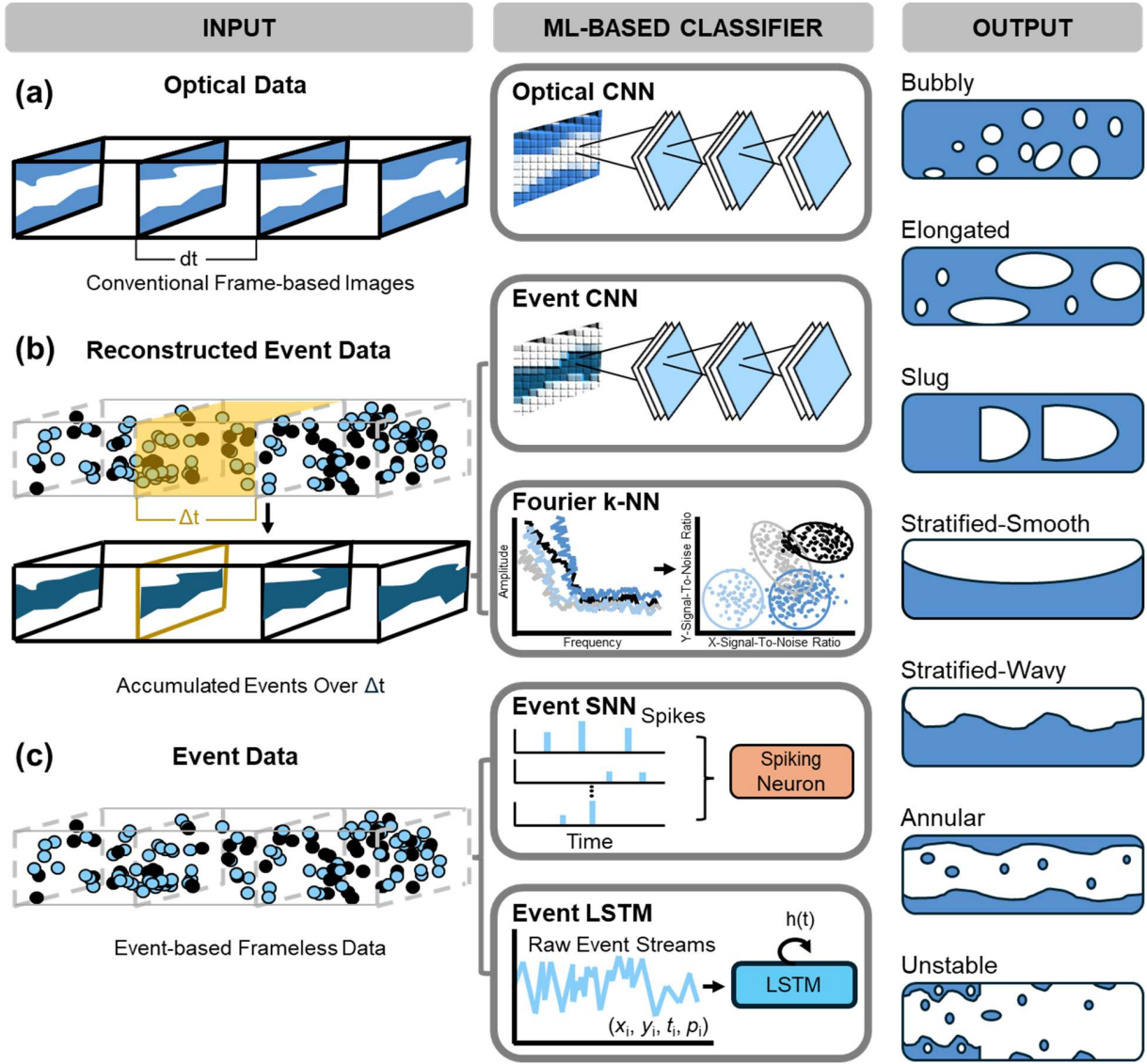

**Figure 3. ML Model Architectures for Flow Classification Task.** Both emulated and raw event data are processed identically, but their preprocessing differs significantly from optical data. (a) Optical data, captured using high-speed cameras, is used exclusively with the CNN, leveraging spatial features from densely packed image frames for classification. (b) Reconstructed event data, generated by accumulating event streams over a fixed time window (Δt), is used by both the *Event-CNN* and the *Fourier k-NN* model. (c) Event data, collected from both real experiments and simulated using V2E, is processed as raw event streams in both the SNN and the LSTM models. These models directly capture temporal dependencies and sequential patterns without the need for event reconstruction, effectively modeling the dynamic transitions of flow regimes.

By integrating optical-based CNNs, event-driven LSTMs and SNNs, and accumulated event frame analysis through *Event CNN* and *Fourier k-NN* models, the proposed framework offers a comprehensive solution for flow boiling classification. Each architecture is designed to align with the characteristics of its input data. CNNs focus on extracting spatial features, LSTMs and SNNs are suited for capturing temporal dynamics, and Fourier k-NN model emphasizes frequency-domain patterns. This multi-model strategy integrates optical and neuromorphic inputs to enable accurate and robust classification across both steady and rapidly evolving boiling regimes. Further implementation details are available in *Supporting Information S2*. Once trained, the models are assessed based on inference time, data processing speed, accuracy, and regime-specific performance, as illustrated in Figure 4.

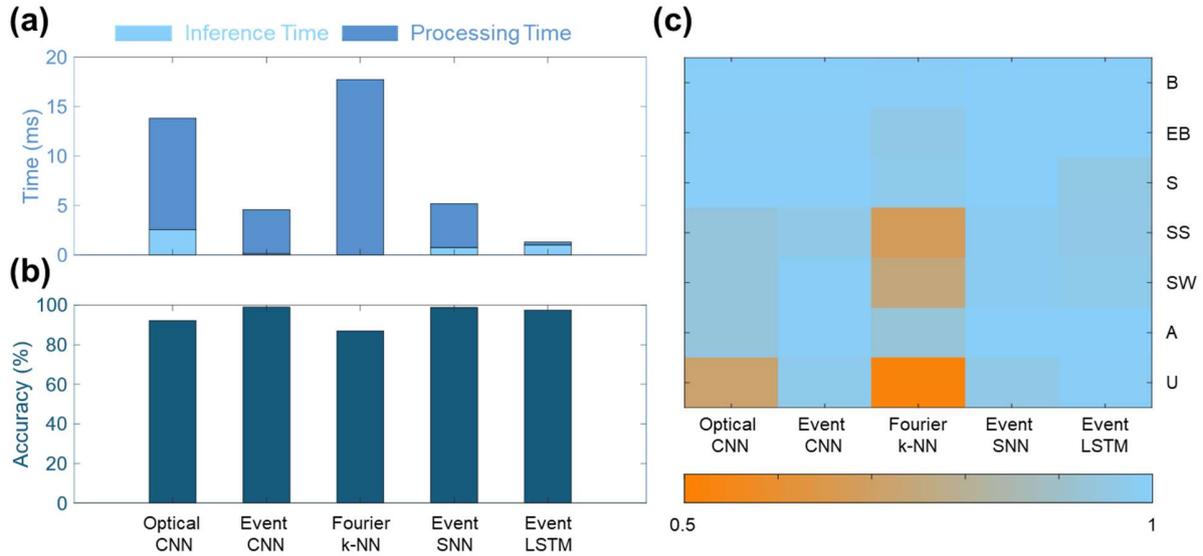

**Figure 4. Model Comparison: Inference Time, Processing Time and Accuracy.** The figure presents a comparative analysis of different model architectures based on their computational efficiency and classification performance. (a) The bar plot compares inference time and processing time across *Optical CNN*, *Event CNN*, *Fourier k-NN*, *Event SNN*, and *Event LSTM*. Event-based models generally demonstrate lower processing times compared to the *Optical CNN*, with *Event LSTM* offering a well-balanced trade-off between efficiency and speed. (b) The bar plot illustrates the classification accuracy of each model, highlighting the superior performance of event-based models in capturing flow dynamics effectively. (c) Classification accuracy heatmap for five models evaluated across seven flow regimes: (B) Bubbly, (EB) Elongated Bubbly, (S) Slug, (SS) Stratified Smooth, (SW) Stratified Wavy, (A) Annular, and (U) Unstable. Accuracy values are represented radially, with higher values extending outward and background rings indicating accuracy bins. Among the models, the *Event LSTM* demonstrates the most balanced performance across all regimes, including unstable flows. Although the *Fourier k-NN* model also uses event data, it exhibits the lowest overall accuracy in complex regimes such as unstable flow. In contrast, other event-based models such as *Event LSTM, Event SNN*, and *Event CNN* show consistently higher performance and robustness, clearly outperforming both *Fourier k-NN* and *Optical CNN* in terms of accuracy and adaptability across all regimes.

*Optical CNN* captures high-resolution spatial features from conventional frame-based images and performs reliably in steady-state regimes. However, it lacks the temporal resolution needed to track rapid transitions, resulting in high inference latency and frequent misclassification of unsteady patterns. Unstable flows present a major challenge due to their transient behavior and overlapping visual characteristics with stratified and annular patterns. As instabilities often emerge from subtle variations in flow conditions, accurately recognizing them demands models capable of capturing both fine-grained spatial details and rapid temporal changes.[52,53] *Fourier k-NN* model achieves the fastest inference by compressing input into compact frequency-domain representations. While effective in stable regimes, it lacks the dynamic expressiveness needed to resolve temporal transitions, resulting in reduced accuracy in flows with overlapping or fluctuating characteristics.

In contrast, event-based models such as *Event CNN*, *Event SNN*, and *Event LSTM*, complete inference and data processing within 5 ms, highlighting their potential for real-time deployment. These models are also highly compact in terms of model size, enabling efficient execution on resource-constrained platforms. *Event CNN* uses event streams into frame-like inputs, enabling powerful convolutional processing and achieving classification accuracy above 99% in most regimes. However, event reconstructions into frames add computational steps which may increase latency and hinder deployment in systems with limited resources. *Event SNN* leverages spike-based encoding that aligns naturally with event-driven data. It performs well in flows with high temporal complexity such as slug and stratified wavy regimes. However, due to the lack of native support for spiking operations on standard hardware, it requires simulation, which increases computational load and latency. *Event LSTM* provides the most balanced performance by operating directly on raw event sequences without reconstruction. It preserves temporal resolution, maintains low computational overhead, and performs robustly across all regimes. Heatmap results and confusion matrices in Figure 4 confirm its superior reliability in detecting unstable and transitional patterns, underscoring its suitability for real-time flow boiling classification.

To investigate these challenges in greater depth, we analyze confusion matrices and additional evaluation metrics that reveal misclassification patterns and highlight how different architectures respond to regime transitions, as presented in *Supporting Information S3*.

**Table 1. Model size comparison**

| Model | Model Size (MB) |
|---|---|
| *Optical CNN* | 2.51 |
| *Event CNN* | 1.27 |
| *Fourier k-NN* | 5.88 |
| *Event SNN* | 0.94 |
| *Event LSTM* | 1.90 |

**Real-Time Classification**

As a next step, a modular software framework is developed to integrate the *Event LSTM* model for real-time classification of two-phase flow regimes. The system is optimized for inference speed, modularity, and reliable data handling under high event-rate conditions.

The event sensor streams data in the format ($x, y, p, t$). Since timestamps do not contribute to classification, they are discarded. Accordingly, inputs to the classifier are ($x, y, p$) only, while dt and Δt are used to align the event sequences with the original optical frames and to define the temporal grouping of events for training. Spatial coordinates are normalized relative to the top-left corner of the flow channel to maintain consistency across different sensor resolutions and placements. To prevent data overload and improve temporal consistency, an event rate limiter restricts the number of events per microsecond, aligning the input with the fixed-length sequence expected by the LSTM.

The classification model, implemented in PyTorch, is designed for real-time use with a compact architecture.

Input vectors of spatial coordinates and polarity are embedded into a 32-dimensional feature space, followed by a two-layer LSTM with 128 hidden units per layer. The output passes through two fully connected layers of 128 units with a dropout rate of 0.3. A softmax layer outputs class probabilities. The confidence shown per window is this softmax probability and thus reflects a window level class belief rather than a physical proportion or a feature detection score. Training is performed on an NVIDIA RTX 3060 GPU using mixed precision, Adam optimizer (learning rate 0.0003), cross-entropy loss, and gradient clipping (max norm 1.0) for 10 epochs with batch size 128.

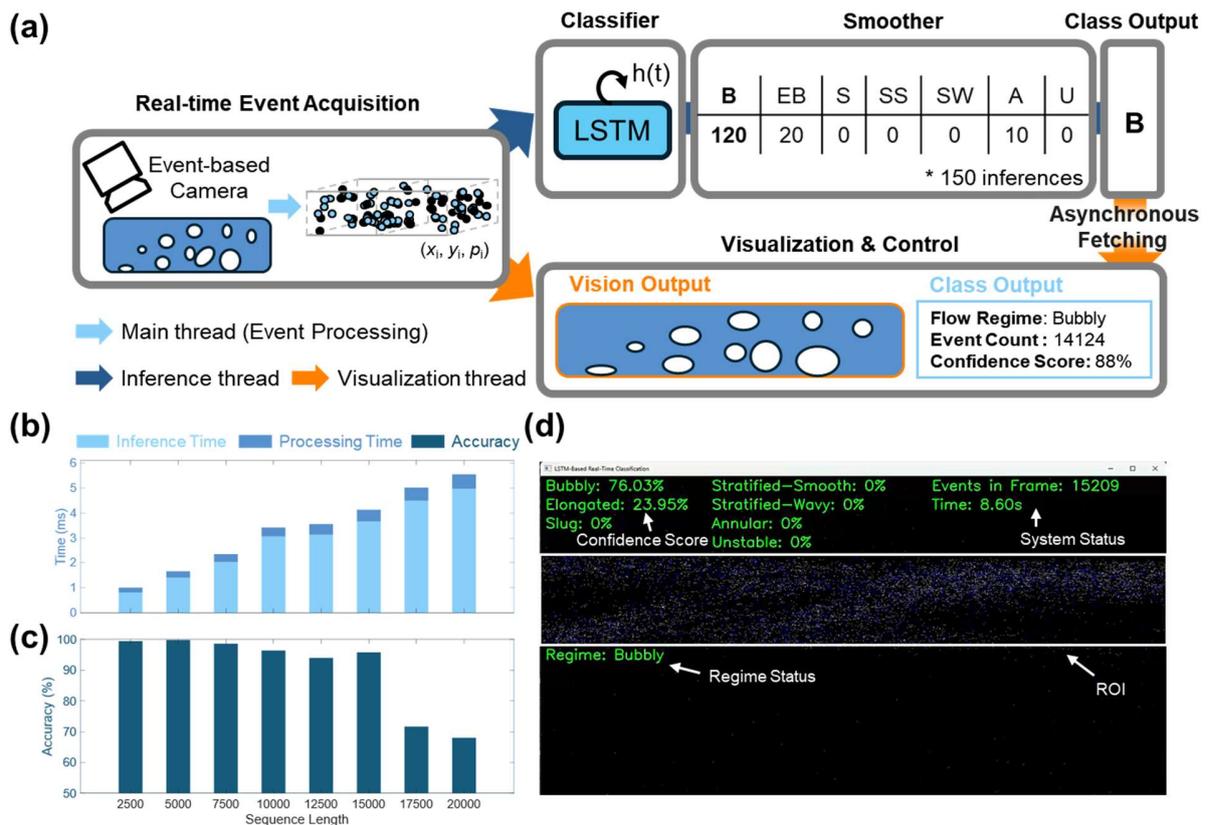

**Figure 5. Real-Time and Online Classification GUI with LSTM**. (a) System architecture for real-time flow regime classification using event-based data. The pipeline begins with system initialization and event acquisition from a neuromorphic sensor with post processing through the main thread. Processed events are queued for inference and visualization. In the inference thread, events are fed to *Event LSTM* model, and outputs are smoothed using a majority vote over the last 150 inferences. The visualization thread displays the most likely classification, event count, and confidence scores, supporting robust monitoring and analysis of flow boiling behavior. (b) Inference time and (c) accuracy comparisons across different input sequence lengths. Longer sequences increase latency but do not always improve accuracy. The best trade-off is observed at 2,500-5,000 events per sequence. (d) The graphical user interface (GUI) shows the classification label, confidence score, system status, and event count. The confidence score represents the proportion of recent classifications matching the current label, enhancing interpretability and decision making in real-time.

As a result, the system runs three concurrent threads for event processing, main, inference, and visualization. The main thread streams event data into a First-In-First-Out iterator, forwarding sequences to a single-item queue. The inference thread processes the latest batch while discarding outdated data to avoid memory overflow and minimize LSTM bottlenecks. A majority voting smoother tracks the last 150 predictions to stabilize outputs. Predictive entropy is reported to reflect model uncertainty and sliding window regime proportions summarize the recent class mix, which together mitigate flicker and separate uncertainty from mixture. The visualization thread renders polarity-coded events with overlaid classifications. The pipeline begins with event camera setup and proceeds asynchronously through thread-safe queues, ensuring robust performance under high event rates and dynamic flow conditions.

We evaluate model performance across different sequence lengths using data. For the analysis in Figure 5b, training and evaluation are limited to the bubbly and elongated bubbly regimes, which exhibit stable event dynamics and sufficient temporal density for sequence-based modeling. Sequence length, defined as the number of raw events per LSTM input, affects both accuracy and inference latency. Longer sequences offer richer temporal context but may increase processing time and introduce noise. Figure 5b shows that a length of 2,500 events achieves the best trade-off, reaching approximately 99.4 % accuracy with 0.81 ms inference time and 0.19 ms processing time. It should be noted that a length of 5,000 events shows a comparable result, reaching approximately 99.7 % accuracy with 1.4 ms inference time and 0.25 ms processing time.

To evaluate real-time performance, the classification system is deployed under actual flow boiling conditions and demonstrates stable, responsive behavior. Although current constraints limit testing to bubbly and elongated bubbly regimes, the system's asynchronous architecture decouples data acquisition, inference, and visualization, enabling continuous throughput and reliable operation with potential for broader applicability.

The model is compact, with a size of 0.95 MB, allowing efficient execution on general-purpose processors. The GUI displays real-time flow regime predictions alongside polarity-coded event visualizations. As shown in Figure 5, the outputs exhibit sufficient temporal consistency for intuitive monitoring. To further enhance temporal coherence, the majority voting smoother filters out short-term fluctuations and effectively reduces output flickering. Demonstration videos are available in *Supporting Information V1, V2 and V3*.

The event-based sensing system also shows strong resilience to environmental variability. Unlike conventional optical methods, the event sensor operates reliably under changing lighting conditions, shadowing, or partial visual obstruction. Preprocessing steps such as spatial normalization and event rate limiting contribute to system stability during extended experiments and rapidly changing flow behavior. Offline tests using V2E-generated synthetic event data incorporating abrupt lighting changes, shadowing, and partial obstructions confirmed stable predictions with our preprocessing and smoother. However, abrupt lighting transitions can still cause spurious event generation, suggesting the need for improved filtering or adaptive bias control in future work.

**Conclusion**

This study presents a real-time classification framework for flow boiling by combining neuromorphic sensing, temporal modeling, and asynchronous software design. Using both image-based or event-based models, we demonstrate sparse, high-temporal-resolution data can be used to accurately and efficiently identify flow regimes. Unlike traditional frame-based systems, which suffer from latency and high processing demands, our approach processes raw event streams directly, enabling millisecond-scale inference while capturing fast and subtle flow transitions.

Among the evaluated models, *Event CNN* delivers high accuracy but incurs longer inference and processing times. *Event SNN* also achieves strong accuracy but requires specialized hardware to operate efficiently. In contrast, *Event LSTM* offers the most balanced solution for real-time deployment. It achieves 97.6% accuracy with just 0.28 ms of processing time and runs directly on raw event data without requiring preprocessing or specialized hardware.

To ensure real-time responsiveness and robustness, the real-time *Event LSTM* further integrates live visualization, adaptive event filtering, and output smoothing to reduce noise. The system remains stable under high event rates and supports intuitive interaction through a graphical interface, enabling accurate monitoring and immediate decision-making in experimental and operational settings.

By achieving reliable, low-latency classification within a flexible and modular architecture, this work shows that event-driven learning systems are well-suited for real-time diagnostics in multiphase flow environments. The methods developed here provide a strong foundation for future advancements in intelligent flow monitoring, control, and prediction across thermal and fluid systems. Performance can be further enhanced by implementing adaptive event filtering that responds to dynamic flow conditions in real time, along with noise suppression techniques and alternative network architectures. These improvements collectively strengthen the potential of event-driven classification systems for both industrial deployment and scientific research. Further directions include integrating neuromorphic processors to enhance the efficiency of spiking neural networks, extending event-based classification to more complex multiphysics domains, and exploring alternative memory models such as state-space networks to improve predictive performance over longer temporal sequences. Future work may also apply these methods to flow condensation to expand the impact of event-based approaches in heat transfer research.

## Methods

### Data Collection

Data collection for flow regime classification in this study leverages localized visualization windows placed within the heated regions of two complementary test sections. One window, centered on a short discretely heated segment of a rectangular channel, captures early-stage boiling such as bubbly and elongated bubbly flow. Another, located midway along a uniformly heated circular tube, records later-stage regimes including slug, stratified, annular, and unstable patterns. Because fully resolving the spatial evolution of boiling along an entire channel would require prohibitively long view lengths that degrade both spatial and temporal resolution, these localized windows provide a practical means to obtain high-resolution footage for accurate regime identification without sacrificing image quality.

The experimental setup in Figure 6a enables steady-state flow boiling under controlled conditions. Flutec PP1 circulates through a closed loop that includes a reservoir, flow meter (Micromotion R-Series Coriolis), and temperature and pressure sensors (Omega PX409) at the inlet and outlet. The fluid is preheated near saturation, then condensed and recirculated after passing through the test section. Figure 6b shows the test section equipped with both a high-speed camera and a neuromorphic event camera to capture flow boiling dynamics. The rectangular polycarbonate channel has a cross-section of 2.5 mm x 5 mm with a total length of 380 mm. A 100 mm heated region is powered by a ceramic heater that provides uniform surface heat flux ranging from 3.7–18.2 kW/m². With the mass flux of 880 kg/m²s, we produce primarily bubbly and elongated bubbly regimes. Classification is based on a local viewing window centered on the heated region. While this allows detailed regime analysis at a fixed point, it does not capture flow development along the full channel. Patterns such as bubbly or elongated bubbly may evolve downstream. Thus, localized classification should be interpreted cautiously, and future work may use wider or multiple synchronized views.

To include the later-stage regimes such as slug, stratified smooth, stratified wavy, annular, and unstable patterns, we leverage additional high-speed optical recordings from a facility with extended boiling capability.[54,55] This closed-loop system uses deionized water and provides access to advanced boiling regimes that are essential for constructing a complete and balanced dataset as shown in Figure 6c and d. The test section uses a 0.25-inch diameter copper tube, 0.955 meters in length, under saturated flow boiling conditions. Uniform heating is provided by conduction from brass blocks on both sides, each made of two C-shaped parts with a 3.5-inch outer diameter, 0.25-inch inner diameter, and 4-inch length. This design ensures strong thermal contact and boundary control, enabling consistent flow pattern generation and allowing clear observation of regime transitions along the channel for insight into the spatial development of boiling behavior. The mass flux ranging from 140–255 kg/m²s with an applied heat flux of 6.8–70 kW/m² yields stratified, slug, annular, and unstable regimes. Detailed operating parameters are summarized in *Supplementary Information S2*.

The optical images obtained are converted to event information using V2E emulation tool.[56] V2E simulates neuromorphic vision sensors by detecting pixel-level brightness changes between consecutive frames and generating events. This process closely mirrors real event cameras, where each pixel triggers asynchronously based on logarithmic intensity changes. The resulting output is a stream of events with spatial and temporal properties similar to actual sensors. This conversion ensures consistent data representation across flow regimes and facilitates integration with neuromorphic machine learning (ML) models. Additional details are provided in *Supporting Information S1*.

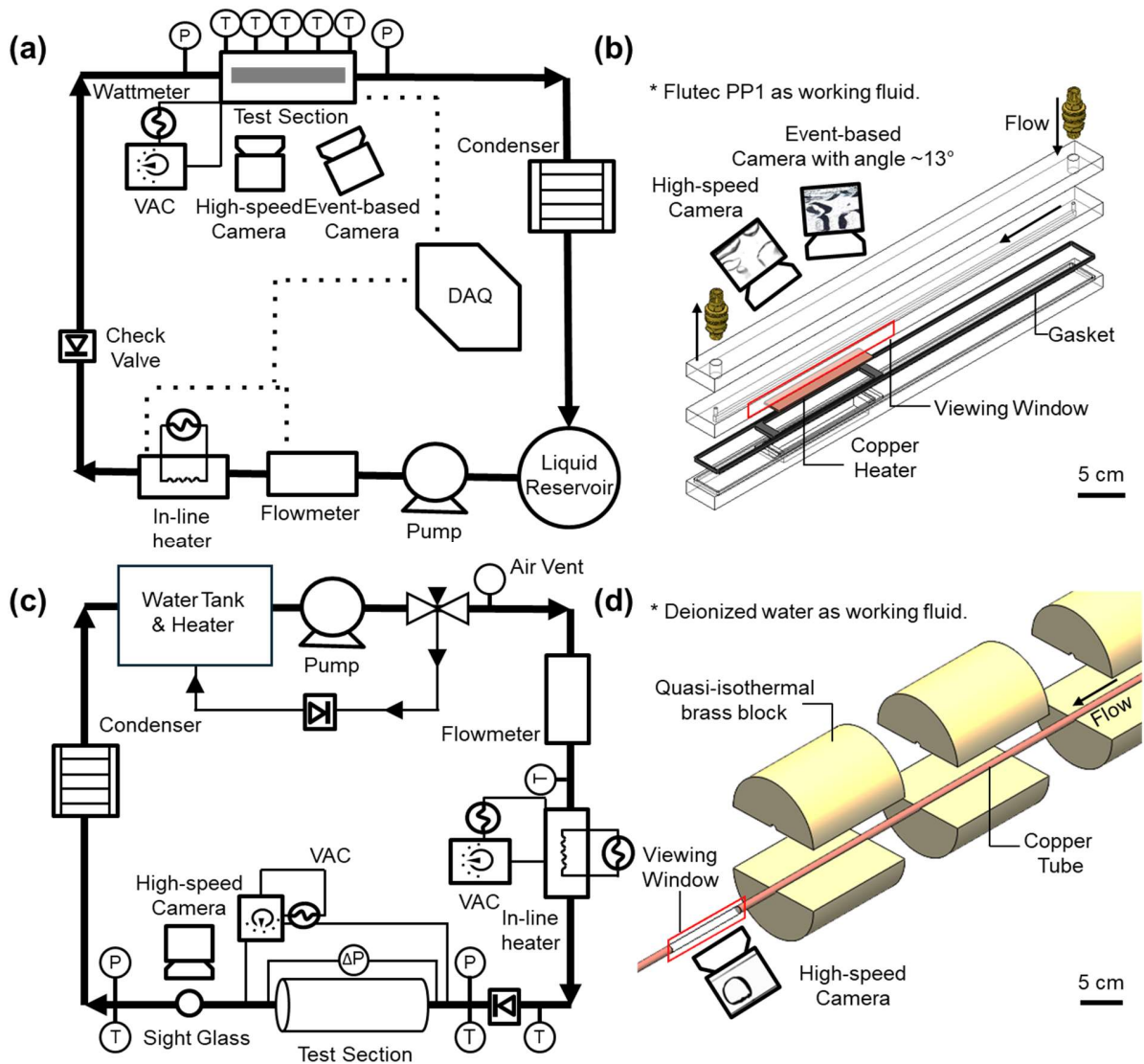

**Figure 6. Visualization For Flow Boiling.** Two flow boiling setups using different working fluids are employed in this study to capture various flow regimes. Flow boiling systems include: (a) a channel with bottom heater and (c) a channel with quasi-isothermal brass blocks —both designed to visualize distinct flow patterns. (b) A detailed view of the polycarbonate test section with bottom heating and optical access, designed for high-

speed and event-based imaging. (d) A schematic of the brass heating assembly, showing C-shaped blocks used to maintain thermal contact and quasi-isothermal boundary conditions along the cylindrical test section.

**ML Data Preparation**

Event-based models (e.g., *Event SNN* or *Event LSTM*) are compared with image-based models (e.g., *Event CNN*). For this, we convert asynchronous event streams into a format compatible with image-based models. Events, both collected and emulated, are grouped by fixed quantities and aggregated into two-dimensional frames. Events are filtered by region of interest and time, then sorted and accumulated until the threshold is reached. Frames are generated by mapping polarities to pixel intensities, optionally separating positive and negative events into different channels. This format preserves temporal dynamics while enabling spatial feature extraction.

The accumulation strategy reduces data redundancy, minimizes memory overhead, and captures meaningful interfacial changes in flow boiling. Event count per frame is optimized per regime to avoid temporal blurring and retain sharp representations. Accumulation time ($\Delta t$) is adapted to flow conditions. High-activity regimes such as (B) and (EB) require shorter accumulation times, whereas more structured regimes like stratified smooth or annular need longer windows due to lower event density to reach the target event count. In parallel, preprocessing pipeline prepares input for models using raw event vectors. Asynchronous event streams are converted from h5 to csv format, keeping spatial coordinates and polarity, while discarding timestamps (*t*). Accordingly, dt, $\Delta t$, and t are used only to construct and synchronize windows and are not provided as explicit model inputs, clarifying that temporal stamps exist in the stream but are excluded from the learning signal. We use fixed length sequences of 5,000 events for time series learning.

This dual preprocessing framework supports both spatial and temporal modeling by generating image-based and event-based datasets from the same raw event streams. Table 2 summarizes the final dataset, which includes real event and synthetic events from the V2E emulator. The dataset spans seven flow regimes, with accompanying information on frame counts and accumulation parameters. Additional details are provided in *Supporting Information S2*.

**Table 2. Summary of Dataset with Flow Regimes**

| Regime | Total Optical Frames | Total Event Frames | Total Accumulation Time (ms) |
|---|---|---|---|
| (B) Bubbly | 6,450 | 8,429 | 1.3 |
| (EB) Elongated Bubbly | 4,105 | 4,835 | 1.2 |
| (S) Slug | 878 | 6,081 | 7.6 |
| (SS) Stratified Smooth | 668 | 3,721 | 14.4 |
| (SW) Stratified Wavy | 1,555 | 10,203 | 6.3 |
| (A) Annular | 2,405 | 18,032 | 7.7 |
| (U) Unstable | 750 | 2,471 | 4.6 |

## Acknowledgements

The authors gratefully acknowledge funding support from the Office of Naval Research (ONR), with Dr. Mark Spector serving as the program officer, under Grant No. N00014-22-1-2063 and MURI Grant No. N00014-24-1-2575.

## Conflict of Interest

The authors declare no conflict of interest.

## Additional Information

Additional supporting information may be found online in the Supporting Information section at the end of this article.